**Title**

Semantics in robotics: environmental data can't yield conventions of human behaviour

**Author**

Dr Jamie Freestone

School of Engineering, Australian National University

jamie.freestone@anu.edu.au

**Abstract**

The word *semantics*, in robotics and AI, has no canonical definition. It usually serves to denote additional data provided to autonomous agents to aid HRI. Most researchers seem, implicitly, to understand that such data cannot simply be extracted from environmental data. I try to make explicit why this is so and argue that so-called *semantics* are best understood as data comprised of *conventions* of human behaviour. This includes labels, most obviously, but also places, ontologies, and affordances. Object affordances are especially problematic because they require not only semantics that are not in the environmental data (conventions of object use) but also an understanding of physics and object combinations that would, if achieved, constitute artificial superintelligence.

# 1 INTRODUCTION

It is difficult to find a canonical definition of *semantics* in robotics.[1] The term is used to describe the kind of data or information that an autonomous system would utilise in human–robot interaction (HRI), namely high-level abstract information such as labels, categories, affordances, places, and so on. This much is unremarkable and there are countless teams working on advances in these areas. But some researchers go further. They envision systems that autonomously understand the larger meaning of objects and events in their environment. They don't need to have the semantics provided, for they will infer or extract it directly from raw data. This kind of semantics is always in the near future. Yet-to-be-invented robots will perceive their environments and, unsupervised, will know the possible affordances of objects, the uses of a room, and the intended meaning of utterances from their human interlocutors. They will do so without being linked to pre-existing databases of high-level information and without exhaustively observing and then imitating humans. The dream, in other words, is to have an artificial agent that can do what we do when we enter a new environment for the first time and instantly comprehend not just the raw sensory data, but the broader meaning and significance of what we sense.

It is only a dream. The problem is that humans also cannot perform this feat, although the way we think can lure us into believing we can.

In this paper I outline a simple idea: *robots cannot extract from the environment data that are not in the environment*. Almost all the data under the banner of "semantics" is of this form and must therefore be provided or learned in earlier forays. A concrete example, taken from the literature, is recognising an ornamental mug in a home (Kollar *et al*. 2013; Sarathy & Scheutz 2016, p.597). It might look indistinguishable from other mugs. Its geometric properties might suggest it can be used as a drinking vessel. But it is not to be used that way. To learn this, the robot would need to be privy to a set of *conventions*: an idiosyncratic history of behavioural regularities of certain humans. This history, though separately learnable, is not in the environment.

I define *environmental data* as any data that could be extracted from the environment. This largely overlaps with *sensory data*. But in practice, such data might also be supplied and hence not obtained directly through sensors. Environmental data covers supplied and sensed data that is, in principle, extractable from the robot's environment.

There are many forms of non-environmental data, one of which is *semantic data* or semantic information, or simply and most commonly in the literature, *semantics*. There are many problems in robotics that employ the concept of semantics: semantic grasping, semantic reasoning, semantic segmentation, semantic place recognition, semantic SLAM, semantic mapping, semantic communication. All these problems require data that are abstract and human-oriented: labels, places, affordances, ontologies, categories, classes, *etc*. These data map environmental data to human actions. Semantics are therefore *conventions of human behaviour relating to aspects of the environment*, *e.g.* the labels humans typically attach to certain objects, or which objects they typically employ for certain affordances. In both examples, the object is part of the environment and amenable to sensing (environmental data). The associated human action related to the object is also part of the environment and can be sensed in any particular instance (*e.g.* video footage of the instances of human behaviour). But because what is important is a history or established convention of behaviour, a system would generally need to observe several trials (in order to learn the convention) or separately be provided with it. In any case, human-free environmental data cannot yield semantics.

---

[1] Definitions of *semantics* are rarely given. Honourable exceptions: Gemignani *et al*. 2016; Kostavelis & Gasteratos 2015, p.87).

Most researchers in robotics, and other fields related to AI, already understand, at least implicitly, this point (*e.g.* Kalfa *et al.* 2021; Russo *et al.* 2021). For them, I aim to make explicit the nature of this error and clarify its implications for future research. For those who advocate for the more ambitious version of semantics, I hope to make clear why it is untenable.

I offer insights from the philosophy of mind. Philosophers have been grappling with an analogue of this error as it occurs in efforts to understand human cognition. My hope is that some useful ideas from philosophical debates over the nature of conventions and semantic information will help engineers avoid some dead ends and allow them to discriminate between work in semantics that is promising and that which is doomed.

In Section 2 I give some examples from the literature and distinguish two versions of semantics in robotics and AI. Section 3 briefly surveys the problem of semantics in other areas. In all cases, the lesson is: a system cannot extract conventional or contextual information from raw data. Neither can humans. In Section 4 I outline why we think we can. The implications for semantics in various problems in robotics are covered in Section 5, along with a subsection on object affordances, which present a special case involving additional philosophical implications.

## 2 SEMANTICS IN ROBOTICS

Certain words jump out at a philosopher of mind while reading robotics papers: *grounding*, *beliefs*, *affordances*, *inference*, *reasoning*, *embodiment* — and *semantics*. Occasionally, definitions are provided. Sometimes they overlap with definitions in philosophy or cognitive science. Typically, the terms are left undefined. This can lend cover to researchers who inflate the abilities of current and future systems. This is not to criticise robotics researchers, who use the prevailing terminology to indicate what others seem to be indicating by those terms. (In other words, they follow *conventions* of use). A word like *semantics* is not being used "incorrectly" just because it is used differently to how philosophers or linguists would use it.[2] The problem, rather, is that certain capacities are being assumed that are impossible and this is partly abetted by the term *semantics*, which is vague enough to refer to highly speculative capacities of future AI systems.

There is also a lack of clarity over what humans can and cannot do, which leads some researchers to project imagined or supposed human capacities onto autonomous systems. The prime example is the hope of autonomously extracting, from environmental data alone, information about how symbols, places, or objects are used by humans. *This kind of semantic capacity is impossible for robots to achieve*. That is not because of current technical limitations. Nor is it because certain human capacities can never be emulated in artificial systems. It is because even humans cannot extract such information because doing so is physically impossible. We effortlessly learn and remember associations between current sense data and previous experience, so it *seems* to us as though we simply extract the semantics from the environment directly (I expand on this in Subsection 4.2). This apparent capacity of humans presumably forms the basis of many overly ambitious hopes for semantics in autonomous systems.

## 2.1 Examples of "semantics" from robotics and AI papers

I begin with an example from place recognition. The hope is to determine what a place is used for — *e.g.* what type of room it is — purely from visual information and without *a priori* knowledge of the new place (note that in all the examples in this section, the italics has been added):

> Therefore, cognitive robots should be able to carry out semantic inferences based on mechanisms of interpretation of the context, *even when places are visited for the first time*. (Crespo *et al*. 2020, p.18)

Similarly, from a highly influential article in the field of place recognition:

> That is, *the robot should be capable of classifying and producing labels for places about which no prior knowledge is available*. In other words, the system should be able to generalize the knowledge gained by exploring a specific spot, so as to *infer about the semantic content of any other similar place*. (Kostavelis & Gasteratos 2015, p.1461)

The same ambition is held for determining object affordances. Researchers hope to one day have systems that can infer an object's possible uses or functions, merely by looking at the object:

---

[2] Note in NLP related topics (like semantic parsing or semantic search) there is a much closer link to semantics *qua* the semantics of language found in linguistics (Nirenburg & Raskin 2004).

> Although the necessity of affordance perception from 3D information recovery, such as optical flow, has been stressed in previous work, we [...] intend to *generalize towards the use of arbitrary features that can be derived from visual information* (Fritz *et al*. 2006, p.4)

> It has been shown in psychology that functionality (affordance) is at least as essential as appearance in object recognition by humans. In computer vision, most previous work on functionality either assumes exactly one functionality for each object, or requires detailed annotation of human poses and objects. In this paper, we propose a *weakly supervised approach to discover all possible object functionalities*. (Yao *et al*. 2013, p.2512)

Beyond the specifics of place recognition and object affordance, some researchers attempt a new paradigm in signal processing or communication. Semantic information, which is somehow latent within the classical or Shannon information of a signal, can be extracted or decoded:

> Semantic (Source) Encoder: detects and extracts semantic content (e.g., meaning) of the source signal and compresses or removes the irrelevant information. (Shi *et al*. 2021, p.46)

The researchers admit, later in the paper, that:

> There is still lacking a simple and general solution for quick semantic information detection and processing that can be implemented in resource-limited devices. (Shi *et al*. 2021, p.47)

The claims boil down to the aim of analysing data from the current scene or environment and somehow inferring or extracting contextual, historical, or otherwise related information that is not physically present in said environmental data.[3] Das and Chernova (2021) ask:

> How can we *autonomously extract contextual information grounded in an environment* to provide meaningful explanations of system failures to everyday users?" (2021, p.3034)

---

[3] These kinds of claims are peppered throughout literature on semantics, especially when future systems are imagined that will do everything current systems can do, but autonomously or unsupervised. Some other examples include Liu *et al*. (2023) on learning what's not in the environmental data; Balaska *et al*. (2020) on unsupervised semantic trajectories; Aotani *et al*. (2017) on "autonomous" deep learning for semantic maps, but with classes that are human supplied; Sarathy & Scheutz (2016) on autonomously learning object affordances, but which can't solve the problem of handing scissors handle-first; and Strinati & Barbarossa (2021) and Xin *et al*. (2024) on outlandish 6G semantic encoders.

They claim to have a method that will,

> *autonomously extract semantic context from novel scenes*, thereby providing detailed explanations even for scenes and objects not previously encountered by the robot. (2021, p.3034)

All these examples assume that wider contextual information relevant to human behaviour is embedded in environmental data, waiting to be discovered. They stand in contrast to the weaker version of semantics, which is about finding efficient ways of linking or associating environmental data with the appropriate (and separate) semantics. Again, I emphasise that the mainstream of researchers working in robotics fall into this latter category, in practice, even if they've never conceptualised semantics in this way.[4]

**2.2 Definitions of semantics**

In the absence of anything like a canonical definition of *semantics* in robotics or allied fields, I identify two implicit definitions and advocate the first one.

1. **Semantics as *conventional***: a conservative, operational usage that frames semantics as high-level, abstract, or auxiliary data, generally for use in HRI. The presumption is that the meaning or significance of the semantics is dependent on conventions of human use or behaviour. For example, labels that are attached to autonomously gathered visual data: in this case, the labels are semantics, because they are entirely contingent upon conventions of human behaviour (*i.e.* language use), which we desire the robot's autonomous information processing to relate to.

2. **Semantics as *contentful***: a more radical, potentially metaphysical claim that certain types of data have inherent or intrinsic properties. This is semantics as the true or accurate meaning of a signal or information, which is extractable from the data itself because it is contained in the data. In short, the data possess *content*, regardless of context, which determines their meaning[5] — *e.g.* if the labels of objects were somehow determined by the object's visual or physical properties.

---

[4] See for example Brohan *et al.* (2023); Huang *et al.* (2022); Russo *et al.* (2020); Russo *et al.* (2021); Zender *et al.* (2008).

[5] In philosophy, linguistics, and cognitive science, *content* is roughly synonymous with *meaning* or *semantic information*. *E.g.* cognitive scientists speak of the *mental content* that characterises thoughts; linguists analyse the *semantic content* of a sentence; collectively these phenomena would be called cases of *representational content* by many philosophers who also talk of the *propositional content* of a proposition or that-statement (Shea 2013; Schulte 2023).

The conventional version is unobjectionable. The researchers involved are attempting what is possible, though technically difficult. What's more, they seem to recognise, perhaps only implicitly, the crucial intuition that one cannot get out of the data more than is in there, that what makes something *semantics* is that it is pregiven or predefined data that can be linked to environmental data. Most notably, they make progress in these endeavours.

The contentful version is dubious. The claims are almost always prospective: the near future will involve a more sophisticated algorithm or information processing paradigm that will somehow extract semantics autonomously and from the given data alone.[6] It is not clear whether the researchers making these claims literally believe them. The claims might be hype or puffery added to the paper to make the current work sound as though it has more future potential (perhaps tellingly, the claims made in abstracts are often bolder than those in the body of the paper). I suspect these ambitions, which are disclosed in well-cited papers and often for prestigious journals, are sincere but misguided. Either way, it is worthwhile demarcating the contentful (strong) and conventional (weak) versions of *semantics* and making explicit why the contentful version is incoherent.

## 3 SEMANTICS IN OTHER DISCIPLINES: CAUTIONARY TALES

Other disciplines study something akin to what engineers call *semantics*. Linguistics, cognitive science, and several branches of philosophy have produced large literatures on questions relating to how meaning is extracted from text, speech, visual information, mental representations, symbols, artworks, memories, and logical propositions. Depending on the discipline and the precise topic or problem, this encompasses notions like *intentionality*, *reference*, *meaning*, *grounding*, *denotation*, *mental content*, *representation*, *signification*, and indeed, *semantics*. One umbrella term for all these phenomena is *aboutness* (Dennett 2017 p.112; Rosenberg 2018, p.120). This captures the sense that some objects, patterns, or processes have some relation to something other than themselves. A rock is not about anything other than itself; but a rock with an icon of a dog carved into it may be interpreted as having some meaning beyond its physical properties as a rock. The icon makes the rock "about dogs," or "represent dogs," or "carry information about dogs," or "contain dog-related content". But explaining how this works is not easy. The problem of aboutness is as old as philosophy (Rosenberg 2013, p.3). The last one hundred and fifty years has seen concerted efforts in Western philosophy to account for aboutness in a formal, scientific, or naturalistic manner. This includes many efforts to formalise semantic information to provide a semantic equivalent to Shannon–Weaver information (Bar-Hillel & Carnap 1953; Floridi 2004; Karnani *et al*. 2009; Nirenburg & Raskin 2004).[7]

---

[6] For example, Crespo *et al.* (2020 pp.15–6) refers to future automatic label detecting work as demonstrated in Crespo *et al.* (2017, p.629), which only gestures at a system that can somehow infer its own labels.

No such attempt has been an uncontroversial success. It is a vast topic and generalisations are bound to be crude, but it is fair to say there is no accepted solution to the problem of semantic information or, more generally, aboutness (Ramsey 2007; Schulte 2023). Some philosophers, observing the lack of progress, have concluded that the relation of aboutness is simply nonexistent or at least sufficiently different to our intuitions of aboutness, that it is worth discarding from scientific and technical explanations (Freestone in press; Quine 1962, p.162; Rorty 1980, 22–7; Rosenberg 2015; Veit 2022). That is, aboutness should be treated as an artefact of our cognition — a feature of how we comprehend thought, language, and symbolic phenomena — and not as a feature of how the world works. I think this is the sanest way forward.[8]

Even without the drastic step of rejecting aboutness, one can take the pragmatic approach advocated by the physicist of information, Ed Fredkin: "The meaning of information is given by the process that interprets it" (2009). The semantic meaning will be receiver-relative and context-dependent as opposed to being determined by the content of the message in isolation. A version of semantics that does not rely on *content* is not only possible but preferable — and indeed it is basically the version of semantics already gestured at by Shannon and implicitly endorsed by many in robotics and AI.

**3.1 Semantics in information theory and communication**
There is a continuity between the more modest, contextual version of semantics and the original framing of semantic information. Claude Shannon's "A Mathematical Theory of Communication" (Shannon 1948), the founding document of information theory, provides a salutary episode in recent history, and one that might appeal to roboticists and engineers. On the first page, Shannon says:

---

[7] Throughout this article I treat Shannon–Weaver information (or simply Shannon information) as the agreed upon standard for the formalisation of information for practical purposes in engineering, computing, robotics, etc. But there are other conceptions of information and there is no agreed upon, unifying account that links, say, Shannon information to Kolmogorov information to von Neumann entropy to Fisher information — although all of them are formally well defined and accepted in their respective domains (Adriaans 2024; Lombardi *et al*. 2014). Philosophers have long debated the status of information, even without considering semantics. I make no claims to such unification. I employ Shannon information, as used in communication and allied fields, because it is familiar to engineers and a nonpareil example of a pragmatic engineer's approach to a conceptually hazardous domain.
[8] The philosophical literature on semantics is vast and tangled and is not really compatible with the semantics as discussed in robotics. It concerns deep philosophical topics like truth conditions, possible worlds, propositional attitudes, *etc*. Most of these concern semantics in language. Because I advocate an eliminativist view of content (Rosenberg 2015; Veit 2022), I see the broader application of semantics to any format or medium of information, as we see in robotics, as a welcome generalisation and deflation of semantics more generally. See Rapaport (2017) for a philosophical work on semantics that is compatible with the semantics in information theory, computer science, and indeed robotics.

> The fundamental problem of communication is that of reproducing at one point either exactly or approximately a message selected at another point. Frequently the messages have *meaning*; that is they refer to or are correlated according to some system with certain physical or conceptual entities. These semantic aspects of communication are irrelevant to the engineering problem. (1948, p.379; italics in original)

Although the statistical or syntactic properties of a message (number of bits, entropy, redundancy) can be measured and quantified by analysing the message itself, its semantic content is an entirely separate question. No one has since developed an equivalent theory for quantifying semantic information.

The key point is Shannon's comment on the "semantic aspects" or "meaning" of a message. He says they are not in the signal itself (as the statistical properties are) but are separate to or beyond the physical instantiation of the message. Messages may "refer to" or be "correlated...with" other "physical or conceptual entities". In other words, they can be associated *by the receiver* with some other objects or events and this makes them meaningful to the receiver. Any analysis of *only* the message — comprised of some data — will fail to illuminate the semantic aspects, including the sender's intended meaning. The semantics depend on how the message has been correlated, in the past, with something else. The receiver of the message must be privy to those correlations, otherwise they will be able to analyse only the statistical or classical informational properties.

It's worth briefly articulating why the intended meaning of the sender is a nonstarter for engineers. [9] There are two main problems. First, this intended meaning cannot be measured as an output, whereas its effect on the receiver can be. Inasmuch as there is disagreement between the intended meaning and the effect, selective processes, natural and artificial, will weed out senders whose messages are malformed and misinterpreted (Skyrms 2010). Senders needn't have intended meanings of any explicit kind, as is the case in signalling systems involving simple machines or single-celled organisms (see Subsection 4.1). The second problem is that when we discuss semantic information outside of a communication problem, there is often no sender at all. We treat the environment as a *source* of information, a kind of pseudo-sender, but the notion of *intended meaning* breaks down. In communicative situations, senders undoubtedly shape their signals in ways that produce desired effects in the receiver. Their efforts are nonrandom. From the space of possible messages via a given channel, they vastly reduce the number by using redundancy and obeying pre-established codes (conventions). But they still cannot insert their meaning directly into the medium of the message (contentful semantics) for it to be "extracted". Rather they rely on the sender being privy to the requisite conventions to interpret effectively; without this additional information the message is mute data.

---

[9] Shannon's 1948 paper was collected with some additional material, including an introductory essay by Weaver, in a book called *The Mathematical Theory of Communication* (1949). Weaver's contribution elaborated on the notion of semantics. He subdivided Shannon's semantics into two problems: (1) the "semantic problem" which involved the sender's intended meaning; and (2) the "effectiveness problem" involving the correlations or associations of the message (1949, p.9). Confusingly, *effectiveness* aligns better with what Shannon called *semantics*. At times Weaver himself slips to a contentful framing when he speaks of some messages being "heavily loaded with meaning" (1949, p.12), which implies that the meaning is *in* the message rather than correlated or associated with it. An even earlier, less thorough, version of information theory was developed in a single paper by Ralph Hartley (1928), also of Bell Labs. Like Shannon, Harltey saw the need to distinguish the "physical" problem of sending a message from the "psychological" problem of interpreting its meaning (1928, pp.537–8). Again, the need to operationalise a notion of information all but forced Hartley to abandon, from the first, an attempt to quantify or detect the semantics of a message.

### 3.2 Comparison with semantics in robotics

The distinction between Shannon information and semantic information is clear in the case of communication. The analogy to robotics is imperfect, but we will use it advisedly.

In communication, semantics are any associated actions taken by the receiver upon receipt of the message. I use *actions* broadly. These could be external behaviours (physical actions) or internal behaviours (updating knowledge, changing posture, *etc*.). A latent assumption in early work on communication was that the sender and receiver are humans, and so the relevant actions are human behaviours. But proposed 6G semantic encoders have a receiver that is an autonomous system which will encode the message in terms of its relevance to human users (Strinati & Barbarossa 2021; Xin *et al*. 2024, p.4). I argue that the proposed encoder will not be able to extract from the signal any such semantics, and even an uninitiated human cannot do this. The semantics — the effects of a message — depend on separate experience. Specifically, one needs to know which messages map to which actions. Without this, even a human is receiving a meaningless message in an unknown tongue; *a fortiori*, an autonomous system cannot know, from the message alone, how it maps to *human* actions or effects. It is the message–action mapping which constitutes the semantics.

In robotics, as opposed to communication, an autonomous agent will take environmental data — the "message", albeit without a sender — and map it to a set of actions, according to a policy, rules, an objective function, etc. But robotics researchers would not regard this as semantics (at least my research has not revealed any who do). Regardless of how sophisticated and abstract the methods for marrying environmental data to behavioural effects might be, they are not considered to be *semantics* (even though they form a kind of message–action mapping) if those actions are solely robot-centric.[10] The *sine qua non* for semantics in robotics is that of *human* effects. The semantics we provide to robots — labels, semantic maps, ontologies for place recognition, object affordances — are derived from a set of effects on *human* agents that we want the robot to be aware of. They are precisely not recoverable from environmental data, because they are about how humans *respond to* certain environmental data of their own. In other words, *they are an imported set of message–action mappings already known to be used by humans*.[11] They can then be used to inform the robot's actions. A robot on an assembly line typically has no need for these. A domestic robot, meanwhile, does need to behave with respect to categories established by and for humans in the past. And because semantics are predefined according to exogenous human needs, they are not discoverable from environmental data: the robot needs separately to learn or be given the relevant human conventions. Here again is the imperfect but useful bridging analogy with communication.

---

[10] Note that even the behavioural conventions of other autonomous agents are not considered semantics. In the literature on swarms and multiagent systems, robots that need to interact with one another (but not humans) generally do so according to coordination protocols or rules for inter-agent interaction, where shared reference frames are established, etc. Again, this kind of sharing of data is not termed *semantics*. Perhaps in the future, with robots from different manufacturers interacting with one another and having to treat one another as human-like agents, we might see *semantics* applied to robot–robot interactions.

> **Communication**. Shannon distinguished between:
> 1. *messages*: the data or signal being transmitted; and
> 2. *semantics*: the message's effects on a human receiver.
>
> **Robotics**. We can distinguish between:
> 1. *environmental data*: gleaned by the robot from sensors or simply provided by programmers; and
> 2. *semantics*: data from human interactions, *i.e.*, historical effects on human receivers of certain data (the semantics in communication, according to Shannon).

This is a way to retain the use of the word *semantics* in robotics and AI — which is surely established — while aligning it with Shannon's definition of semantics in communication. In both areas, semantics is distinguished from raw or environmental data, which can be processed independently of human conventions. Those conventions of human actions in response to certain data are semantics, which can be framed informally as the *meaning* people derive from certain information.

### 3.3 Semantics in other sciences

---

[11] Again, these definitions don't fully align with semantics in philosophy, computer science, linguistics, etc., which differ among themselves. The best discussion of semantics, general enough to apply across these disciplines, is Rapaport's (2017; 2018). He says it is "syntax all the way down" but one can get semantics out of the relations between two sets of elements whose internal relations are all pure syntax (2018, p.228). A set comprised of the two sets and their relations is still itself a set of syntactic information. But it also comprises semantic information in the Shannon sense, or the sense advocated here, namely that it is formed from some *mapping* of environmental data (syntax) to data describing human responses (semantics). In Rapaport's scheme, such data needn't be related to human behaviour. Robots would already have semantics in this sense. But, following robotics researchers, I restrict *semantics* to that based on human behaviours; and hence I align it with the concept of conventions, which are essentially observable regularities in human behaviour.

Other branches of science encounter the same difficulties in demonstrating a contentful version of semantics. In Shannon's wake, Wiener (1967) was one of the first to try to transplant information theory from communication to living systems. Wiener defined semantic information as that which activates something in the receiver, helping it to act effectively (Oyama 1985, p.66). This is concordant with the operationalised definitions found in robotics and related areas. It is essentially the effectiveness aspect highlighted in early communication theory (Shannon and Weaver 1949, p.25, p.75) applied to living and nonliving systems, under the aegis of cybernetics (Wiener 1967).

In biology, there was much debate about whether a gene "carries information about" the environment and how much it specifies the organism's development. There too, there is an approach that emphasises context over content. Genetic information is "meaningless" without the cellular machinery for transcription and translation and the constraints imposed by the extra-cellular environment (Dennett 1995, pp.195–9; Oyama 1985). And in biological communication, communication in nonhuman species is perhaps more obviously pragmatic and use-based than in the sophisticated and open-ended realm of natural language. Sending a message — a chemical signal between cells, a bird call — is simply performing a particular action, part of a behavioural repertoire, and so it is less suggestive to suppose the messages are contentful. One early and influential model in the literature was that of Dawkins and Krebs (1984). They frame an animal signal as a way for one organism to exploit another's muscle power via their sense organs (1984, pp.381–2). Dawkins and Krebs acknowledge the power of Shannon information, but they too question trying to quantify semantic information: "we suggested that it might be better to abandon the concept of semantic information altogether in discussions of the ritualisation of signals" (1984, p.397).

Physicists have also attempted to naturalise or mathematise semantic information. A recent flurry (Kolchinsky & Wolpert 2018; Sowinsky *et al*. 2023) follows work by Rovelli (2018). These approaches look at the correlated information between an organism and its environment. They identify that portion which increases the organism's viability as semantic information (Kolchinsky & Wolpert 2018). Although this is at least a formalised definition of semantic information, it is still receiver- or organism-relative; or, more precisely, relative to the organism–environment relation, that determines which information contributes to viability (2018, p.12). A relational version of semantics does not, from my perspective, run afoul of any philosophical concerns. But it is another example of how concrete attempts to operationalise or formalise semantic information invariably encounter the eye of the beholder problem: one set of Shannon information will affect receivers differently depending on their history — in this case, the organism's evolutionary and ecological history that shaped its phenotype and behaviour. And it is these variable effects of the same (or similar) environmental data on different agents that forms the social-interactive nature of human semantics, *i.e.* ways that humans typically respond to a given context.

## 4 INSIGHTS FROM PHILOSOPHY OF MIND

The key point of this paper is that the semantics of words, like any other information arising from social interaction, is based in *conventions of use*, as opposed to intrinsic properties of the medium.[12] This is another way of saying: context over content.

### 4.1 Convention

Many phenomena have been modelled as conventions: language, money, road rules, etiquette, fashion, *etc*. To understand systems like these, one must be aware of at least some of the history of interactions among participating agents. This is partly because the conventions could be otherwise: a place that drives on the right *might have* adopted a convention of driving on the left instead. Alternatives needn't be completely arbitrary. There might be systemic reasons why some alternatives are more or less likely to be adopted; it is unlikely that a place would adopt driving down the middle of the road as a convention. But without knowing which convention has actually prevailed, one cannot know *ex ante* which convention agents will adopt. It is also necessary to know the history of interactions because that is what constitutes a convention, rather than the present instance or a one-off event.

The first rigorous study of conventions was Lewis' monograph, *Convention* (1969). He analysed the development of conventions via game theory, specifically coordination problems. In such problems, agents have a mutual interest in following the same convention, *e.g.*, both driving on the left-hand side of the road to avoid collisions. The Nash equilibrium in these games is when agents match others' behaviour. Defecting can only be detrimental. Defection must be possible, though, for it to be a convention, as opposed to some other regularity where no choice is involved (Lewis 1969, pp.69–70). In the case of the semantics of place, for instance, agents might evince regularities in behaviour in a certain room. Only some regularities will be conventional because only some involved choice. Obeying the law of gravity is not a convention; removing one's hat inside is conventional.

The choice doesn't have to be conscious or deliberate. Perhaps the most important extension to Lewis' work is Skyrms' *Signals* (2010). Skyrms introduces simple signalling games that model senders' and receivers' behaviour in establishing conventions. The agents in many of Skyrms' games aren't conscious or even especially agential. He models the signalling between bacteria and even between cells in the same organism (2010, pp.29–31, pp.118–20, pp.151–3). Indeed, Skyrms' account is evolutionary. He shows how, in just a few iterations, a convention will evolve in a completely blind manner, often depending on chance differences in initial conditions.[13]

---

[12] Note that in the literature on conventions these are often called social conventions to distinguish them from other phenomena that might, in ordinary parlance, be called "conventions". I use the shortened term convention throughout this paper to mean "a regularity widely observed by some group of agents"— which is a philosopher's definition of social convention (Rescorla 2024).

[13] See Skyrms (2010, pp.7–8, p.46, p.64). Note that Skyrms does include an account of content (2010, pp.40–2), though not intentionality (2010, p.42). Content for him is based in how much a signal moves

Conventions offer a powerful and baggage-free way to understand how agents attribute meaning to objects and events with which they *and others* interact. Such attribution does not require the objects or events to possess intrinsic meaning or contentfulness. It does not require that the agent's internal states possess intrinsic meaning. All that is needed is some behavioural regularity, not internal representations and therefore not even an intended meaning (see Grim *et al*. 2004a; Grim *et al*. 2004b; Planer & Godfrey-Smith 2021). Because of this, one cannot second-guess the meaning of some object or event without knowing of other agents' responses to it; the whole point of communication and many other social interactions is to base one's behaviour on other agents' behaviour. All that is required for conventions to develop is that multiple agents interact with the same objects or events and derive some benefit from aligning their responses. The incentive structure native to communication and cooperation will nudge agents towards alignment. Certainly, it will be possible one day to create robots that learn semantics by interacting with humans to alight on the same conventions. Before then, the conventions need to be supplied because they are not available in environmental data.

**4.2 Why contentful semantics are alluring**

The more ambitious hopes for semantics in robotics and AI may stem from overestimating what humans do. When encountering a scene, it might seem like people "extract" semantics from their sensory data. The temptation is to emulate this ability in autonomous agents despite it being impossible for humans. Briefly, here are three factors, courtesy of philosophy of mind and cognitive science, which explain why it *appears* that we extract semantics from environmental data.

**Projection**. (Also *projectivism*, *projectionism*, *the problem of perception*.) The Scottish philosopher David Hume is normally credited as the originator of this idea: "the mind has a great propensity to spread itself on external objects, and to conjoin with them any internal impressions, which they occasion" (Hume 1739/2000, p.112). Human perception works in such a way as to project the associations and conceptual information triggered by an object of perception *onto* the object of perception. One sees a dog and various associations or "semantics" related to dogs are activated in one's mind: crudely put, one's DOG concept is activated. But this concept includes information not at all present in the current sensory data (Dennett 2017, pp.354–8). The perception of the dog, drawing on past experience, is in many ways richer than what is in the data — a classic case of semantics being used to supplement environmental data. But from a person's own perspective, the richer properties of the DOG concept

---

one's probabilities and in what direction — not unlike some of the approaches mentioned in Subsection 3.3. Again, this is a big improvement on content as an intrinsic property (like intentionality). I don't endorse it here, although it may be fruitful as a future theory of semantic information. In any case, Skyrms' content depends, unlike Shannon information, on more than the environmental data itself, *i.e* the content of the signal depends on one's pre-held beliefs (separate, previously obtained data).

seem to inhere in the actual dog in the environment currently being perceived. Some of these properties, such as the dog's colour, are entirely artefacts of the perceptual system. Others could be said be "real" properties possessed by the dog independent of our perception, but which are not actually recoverable from the present instance, *e.g.* the dog's cuteness, whether or not it is a dangerous animal, if it belongs inside or outside, that it can be petted, and so on. These might be properties gleaned from past interactions with other objects in the category of DOG or from past interactions with other humans relevant to this particular dog (conventions).

**Prediction**. These projections are part of a *prediction* of what one is sensing. Cognition is prediction-heavy. An accounting of the brain's traffic shows there are more efferent signals than afferent ones (Dennett 2017, p.169). A fair generalisation is that the last thirty years of cognitive science has seen a shift to emphasising the predictive nature of cognition. Predictive processing or predictive coding, the Bayesian Brain hypothesis, active inference models, and the free energy principle all approach cognition as being a multi-level process of forming predictions to aid behaviour and having those predictions modified by feedback from the senses (Hohwy 2018). Even with highly evolved sense organs, environmental data is often too noisy and/or too incomplete to be relied on: better to learn regularities in the environment and then make informed predictions of what is causing a current perception. Sensory data is then used mainly as an error signal to modify the prediction which was based on past experience (Dennett 2017, pp.167–70). Again, perception is often richer than the current data allow.

**Nonconscious cognition**. A final insight from philosophy concerns the larger project of trying to emulate human capacities in robots. Take, for example, cognitive maps. Humans, like other mammals, have sophisticated systems for navigating space: grid cells, head-direction cells, boundary cells, and so on, which, combined, make up a cognitive map (Rosenberg 2018, 131–8). Our conscious experience of navigating in no way resembles the nonconscious cognitive map architecture. We aren't conscious of space being divided into tessellated triangles forming hexagons, or of grid cells activating a preplay sequence as we take the door to the living room rather than the kitchen (Rosenberg 2018, pp.141–56). Using introspection alone, we could never have obtained the functional details of the cognitive map architecture. Some teams have had great success adapting the nonconscious features of the brain's cognitive maps to perform navigation in robots (Kuipers 2000; Milford *et al*. 2004). But building a system inspired by our *intuitions* of how we navigate space would be a bad or at least unhelpful idea.

All these contribute to the strange situation whereby we expect robots to do something impossible. Robots cannot extract semantics from environmental data; the semantics are in the relations between environmental data and human activity; and these relations constitute separate data. In a sense, robots don't make this error. They process what they can from environmental data. It is we who then expect them to mine that data for the insights we get when we perceive our environment (thanks to projection, prediction, and nonconscious cognition). Forgetting Shannon's dictum, we look for the correlations of some data in the data themselves.

## 5 IMPLICATIONS FOR PROBLEMS IN ROBOTICS

We can classify some of the problems in robotics already mentioned, according to what kind of data they need to be solved: environmental data, semantics, or a combination.

**Labels** are purely conventional: they are solely a product of a history of human interaction. Therefore, they can be provided to the system, as in a database of labels and their associations with environmental features; or they can be learned, as in the case of language models that can accept new vocabulary given by humans. They are the most straightforward case of an autonomous system acquiring semantics to aid in tasks that involve interacting with convention-using agents, namely humans.

**Grasping** may appear to be a purely environmental problem, where the system must comprehend the geometry and physical features of the object to be grasped. However, as in the case of what is called "semantic grasping", contextual or semantic information is sometimes needed (Murali *et al*. 2021; Tremblay *et al*. 2019). As Liu *et al*. (2020, p.2550) suggest, we might like a robot to not only select a secure grasp for an object like a pair of scissors, but to know to pick them up blade-first so they can be safely handed to humans. That kind of information is conventional and so cannot be inferred from the environmental data alone; notably that particular example remains unsolved by Liu *et al*.'s system (2020, p.2553). Grasping in a domestic context is therefore a mixed data problem.

**Place recognition**. Most work in place recognition tries to marry geometric properties obtained from environmental data with high-level concepts, to establish, for example, the type of room the robot is in. This is a mixed problem. Most approaches use ontologies of rooms and objects (Crespo *et al*. 2020, p.2; Pronobis & Jensfelt 2012, p.3515). The ontologies can be provided, constituting a ready example of semantics as a form of information given to aid HRI, not unlike labels. Place recognition is complicated, however, when researchers want the robot to autonomously infer place when they "visit for the first time" (Crespo 2020, p.16). This may require additional semantics if it is a nontypical room. Whereas labels tend to be conserved across populations of human users (from the same linguistic community), place designations are more variable. Examples are manifold. The presence of a bed might lead the robot probabilistically to conclude they're in a bedroom, while they're actually in a studio apartment; or perhaps they are in the garage where an excess bed is being stored before it is sold; or perhaps the room they're in is really the "study" and is referred to as such by the human inhabitants, but occasionally doubles as a guest bedroom. Such cases illustrate that *place* is sometimes determined more by the human behaviours performed in that place rather than the objects contained in it. What's more, there is not any correct answer beyond what the local humans converge on, and, in the case of domestic places, "local" means the particular inhabitants of that home. Additional semantics based on idiosyncratic human conventions would be required.

The other major robotics problem involving semantics is object **affordances**, which is also a mixed problem but raises deeper epistemological issues, necessitating its own section.

## 5.1 Affordances

In robotics, *affordance* tends to be a synonym for *function*, or even *use* (Jamone *et al*. 2018). Object affordances can be highly specific and entail multi-object actions: *cut_with*, *pour_into*, *change_substance_temperature*, and so on (*cf*. Beßler *et al*. 2020; Henlein *et al*. 2023; Ramirez-Amaro *et al*. 2017). This is broader than the robo-centric affordances in early papers on object affordance in robotics, like *liftability* or *stackability* (Jamone *et al*. 2018). And it is much broader than the original concept of affordances outlined in the ecological psychology of Eleanor and James Gibson (Gibson 1977; Gibson *et al*. 1978).[14] Continuing the approach to semantics and conventions advocated in this paper, I use the term *affordance* pragmatically, in line with how it is actually used in robotics and AI papers — and that, fittingly, is in terms of an object's actual use by the relevant humans.

More important than the definition of *affordance* or *function* is what kind of data robots need to use objects. There is a prominent assumption among the more ambitious researchers, that object affordances (*qua* the actual uses) are latent within the physical properties of the object and therefore can be autonomously extracted (Fritz *et al*. 2006; Kollar *et al*. 2013; Liu *et al*. 2020), even though they often find it can't be done yet (Thosar *et al*. 2021, p.6). Sometimes, researchers even claim to extract *all possible* object affordances from environmental data (Yao *et al*. 2013, p.2512).

This framing of object affordances is the contentful version of semantics applied to the functions of objects. If the functions or affordances are "intrinsic properties" (Varadarajan & Vincze 2012) of objects that are somehow inside them, ready to be extracted, then this approach would work. But the affordances that roboticists are interested in are conventional. Most researchers, in practice, recognise at least that affordance detection is limited by the degree of *a priori* knowledge provided (*cf*. Ardon *et al*. 2019; Do *et al*. 2018; Sarathy *et al*. 2018). Indeed, the example of the ornamental mug is telling: the use or non-use of this object is less to do with its physical properties and entirely tied up with local human custom. Just as unfamiliar human customs can be learned by ethnographers embedded in foreign cultures, a robot could one day attempt the "ethnographer's challenge" (Table 1) of learning human conventions of "correct" object use.

---

[14] In ecological psychology, affordances are an object's potentials for action most clearly suggested to a human or nonhuman animal; different species perceive different affordances. For humans (mainly but not exclusively) affordances are suggested not only by natural objects but also artefacts. Object affordances and functions raise a host of philosophical questions about natural versus human-made objects; whether affordances are objective (an intrinsic feature of the object), subjective (organism-relative), or relational (a product of the object's properties and the organism's interests or abilities); and whether a human-made object's function or affordance is determined by the maker's intent, history of use, or current use (Cosentino 2021; Preston 2022; Wilkinson and Chemero 2024). Such issues cannot be tackled here, but the approach I take is that, for a robot's purposes, an object's affordance(s) equals the object's actual use(s) by the relevant community of humans.

The flipside is that an "incorrect" but novel use of an object may violate convention while still performing a task. By convention, a large book might make an ersatz doorstop. A MacGyver-esque robot able to recognise certain physical properties of objects might find it is possible, though unconventional, to use other weighty items for this purpose: a bag of rice, a tub of moisturiser, an urn containing grandma's ashes. Such improvisation to satisfy success conditions would, outside of emergency survival situations, likely annoy its human interactants. (The "survivalist's challenge" in Table 1.)

A further problem is that most human affordances are multi-object combinations. This is an overlooked point in the literature on artefact function, the literature on affordances in humans and other animals, and the literature on affordances in robotics.[15] In nonhuman animals, it may be more common to see relatively isolated object use: a branch is a perch, a seedpod is food, etc. But even simple household affordances described in the robotics are actually part of multi-object tasks. Objects like *cups* are filled with other objects like *juice*; affordances like *cut_with* depend not only on properties of the cutting object (sharp edge, etc.) but also the object to be cut (solid, frangible, etc.). What is needed is complex data regarding multiple objects' conventional uses in combination.[16] Again, humans evidently learn many (but nowhere near all) of these combinatory affordances, often by imitation, sometimes through teaching.[17] But the space of human affordances is vast and is known by humanity as a whole, not any individual. Consider, for example, the set of affordances presented by some common kitchen utensils and household ingredients: thousands of recipes. And this is a tiny subset of all the affordances — embodied in tools, games, farming, war, production, household labour, etc. — in the human repertoire.[18] The hope for autonomous affordance discovery implies an object's relations to other objects are inbuilt. A future domestic robot might be given the procedure for executing certain recipes or might even learn through imitation. But it is fanciful to think a robot might enter a kitchen and "extract" — from the physical properties of eggs, caster sugar, a whisk, and an oven — a recipe for a meringue, as though the potential destiny of materials were contained within them like some *telos* or *kami*.

---

[15] For a partial exception see Moldovan *et al*. 2017. Hassanin *et al*. 2021 offer the fullest discussion of the manifold difficulties of affordance/function detection.

[16] In fact, even single-object affordances are irremediably semantic. Once an object is used it implies a user. This user is, in some sense, another object and so the use entails at least a two-object system. With purely environmental data, an object can be demarcated or isolated and its properties detected. As soon as use or function is invoked, the system whose properties one is interested in expands to become the object plus its user.

[17] See Henrich (2016) on imitation as the driving force behind human cultural evolution.

[18] Ethologists sometimes catalogue a species' behavioural repertoire in an *ethogram*. For many species, the ethogram might run to a few dozen items, with almost all observed behaviour fitting into these categories (Brockmann 1994). No such ethogram for *Homo sapiens* could be produced. Perhaps in our very early history we might have had a behavioural repertoire closer to that of other social primates, like chimpanzees. But such is the open-endedness of cultural evolution — including constant technological and economic innovations — that the human ethogram now increases daily. These in combination breed new affordances still. The growth of affordances, and hence human behaviours, is exponential.

Inferring affordances is perhaps another example of projection. A human, given a new object, sans context, will struggle to guess what it is for. Call it "the archaeologist's challenge" (see Table 1). Without surviving users of the object, archaeologists rely on other objects found in the vicinity to assemble as much context as possible, to then match with potential affordances they already know of. Without any context, the task becomes all but impossible (Preston 2009). How, for instance, could one know an object's actual use was purely ornamental?

A system that would select new combinations of objects *and* infer new affordances faces a greater challenge still. In fact, it is *the* problem of epistemology: that of scientific discovery, creative leaps, and the invention of tools and technologies. To autonomously infer an object's *possible* uses — including how it could be combined with other objects — the system would need to know the physics and chemistry of the rest of the world and to reason about how these might *afford* new uses for objects in relation to the needs of humans or other robots. Call this "the inventor's challenge". Once again, humans cannot reliably invent new affordances and tend to become fixed on the single function of an object they have learned through teaching (German & Barrett 2005). A system that solved the inventor's challenge would certainly be wasted on figuring out how to use cutlery. It would constitute an intelligence that could leapfrog the most intense and concerted efforts of human scientists and engineers. The researchers working towards this may not realise it but they are attempting to build an artificial superintelligence.

I encourage engineers to adopt the conventionalist account for object affordances or artefact function. For roboticists working on current day systems to be deployed in homes or offices, their approach to object affordances should be to tackle the ethnographer's challenge; the archaeologist's and inventor's challenges are impossible in the near term; the survivalist's is pointless.

**TABLE 1: learning object affordances**

|  | *Object(s) provided.* (Follow conventional use, *i.e.* semantics.) | *Object(s) to be selected.* (Identify new object(s) for affordance, ∴ not reliant on semantics/conventions.) |
|---|---|---|
| *Old affordances* (Match object properties to known affordances.) | **The ethnographer's challenge.** Learn and copy the prevailing ("right") conventional uses of objects: <ul><li>to *spear_food* use *Fork*</li><li>for *convey_to_mouth* use *Fork*.</li></ul> Can be provided as semantics or learned in HRI. | **The survivalist's challenge.** Possible but *un*conventional ("wrong") uses; a novel object for an old affordance: <ul><li>fork used for *comb_hair*</li><li>sharp sticks for *spear_food*.</li></ul> An advanced system could discover these. Questionable value outside of emergencies. |
| *New affordances* (Given the object(s), what affordances can be discovered?) | **The archaeologist's challenge.** When the conventional affordance is unknown via lack of context: <ul><li>fork might be solely used for *eat_oyster*</li><li>fork could be used for *whisk_food* along with other objects to make meringue.</li></ul> Multi-object affordances, especially, are open-ended, so one needs to be given the history of use: semantics. | **The inventor's challenge.** The problem of science, innovation, or creativity in general: new uses for novel selections of objects: <ul><li>fork used along with other objects for a new contraption or new game.</li></ul> A system that enumerated the *possible* uses of a combination of objects, on sight, would be tantamount to an ASI. |

*N.B.* A system can infer some negative affordances, *e.g.* a fork can't be used as a liquid, etc. The table summarises positive knowledge of affordances. And the inventor's full challenge would also include creating new objects, something not even mooted yet in robotics literature.

## 6 CONCLUSION

Autonomous systems should, in principle, be able to learn anything conventional. This includes labels, places, and conventional object affordances. Like humans, however, they cannot extract conventions from environmental data alone. Most AI researchers realise this at least implicitly — some don't. This paper hopefully makes explicit the difference between environmental data and *semantics*, which should refer only to data relevant to human actions and therefore HRI.

  As much as one can generalise about a whole set of disciplines, engineers are fairly pragmatic in every sense of the word. They are interested in designing systems that work, which means they're less inclined to ponder, or make, large metaphysical claims beyond practical considerations. They are interested in what can be *done*. Hence, they often approach information from a purely technical or performance-based perspective, *i.e.* Shannon's approach. My sense from many years of reading on this issue — and from my interactions with engineers working in robotics, control, and autonomous systems — is that the *meaning* of information is usually taken to be contextual and based on usage: pragmatics. Semantics are usually assumed to be nothing more than supplementary, pre-existing, or contextual information provided to a system to aid performance in HRI. My admiration for engineers is genuine. In part this is because their pragmatic stance on these matters is, I think, saner than the stance taken by many of us in philosophy, linguistics, or cognitive science who hold out for a science of semantics on the same footing as Shannon's theory.

Recently, a growing number of researchers in AI, ML, and robotics expect their systems to one day autonomously uncover the semantics of their environments. This is a departure from the engineer's pragmatic ethos. My hope for this paper is that the concepts discussed here might help some engineers in these areas to better sift out the more fanciful claims regarding semantics.